%% file: main.tex
\documentclass[journal]{IEEEtran} 
\IEEEoverridecommandlockouts

\usepackage{amsmath,amssymb,amsfonts}
\usepackage{caption}
\usepackage{subcaption}
\usepackage{subfloat}
\usepackage{ulem}
\usepackage[hidelinks]{hyperref}
\usepackage{cleveref}
\usepackage{multirow}
\usepackage[ruled,vlined, linesnumbered]{algorithm2e}
\usepackage{graphicx}

\usepackage[noend]{algpseudocode}
\usepackage{textcomp}
\usepackage{xcolor}

\usepackage{float}
\usepackage{longtable}
\usepackage{wrapfig}
\usepackage{tikz}
\usepackage{standalone}
\usepackage{pgf}
\usepackage{nicefrac}

\usepackage{xspace}
\newcommand{\AMELS}{\texttt{AMELS}\xspace}
\usepackage{enumitem}
\usepackage{nicefrac}
\usepackage{tikz}
\usetikzlibrary{positioning,calc}

\input{colors}
\input{macros}

\renewcommand{\emph}[1]{\textit{#1}}
\usepackage{cite}
\usepackage{booktabs}
\begin{document}

\title{
\texorpdfstring{
Algebraic Multigrid Acceleration for\\ Efficient Label Spreading}
{Algebraic Multigrid Acceleration for Efficient Label Spreading}
} 

\author{
\IEEEauthorblockN{Antonia van Betteray$^{\ast,1}$\thanks{$^\ast$ Equal contribution.}},
\IEEEauthorblockN{Jonathan Klees$^{\ast,1}$},
\IEEEauthorblockN{Miriam Schäfers$^{\ast,2}$}, 
\IEEEauthorblockN{Matthias Rottmann$^1$}
\\
\IEEEauthorblockA{$^1$\textit{Osnabrück University, Institute for Computer Science, Osnabrück, Germany}
\\ \{antonia.vanbetteray, jonathan.klees, matthias.rottmann\}@uni-osnabrueck.de 
}
\\ \IEEEauthorblockA{$^2$\textit{Ruhr University Bochum, Institute for Computer Science, Bochum, Germany}
\\
m.schaefers@rub.de}
}




\maketitle

\begin{abstract}
\input{00_abstract}

\end{abstract}


\setlength{\parindent}{0pt}
\setlength{\parskip}{0.5em}

\section{Introduction}
\input{01_introduction}

\section{Related Work}\label{sec:related_work}
\input{02_related_work}
\section{Preliminaries}\label{sec: Preliminaries}
\input{03_preliminaries}
\section{Method}
\input{04_method}

\section{Numerical Results}
\input{05_results}

\section{Conclusion}\label{sec:conclusion}
\input{06_conclusion}

\clearpage
\section*{Acknowledgements}

This work is based on the Master's thesis of M.S. \cite{Ackermann2026}.

\noindent
J.K.\ and M.R.\ acknowledge support by the German Federal Ministry of Research, Technology and Space (BMFTR) within the project RELiABEL (grant no.\ 16IS24019B).

\bibliographystyle{IEEEtran}
\bibliography{literature}


\renewcommand{\thefigure}{A.\arabic{figure}} 
\renewcommand{\thetable}{A.\arabic{table}}   
\setcounter{figure}{0}                        
\setcounter{table}{0}                         

\clearpage
\appendix
\label{sec:Appendix}
\input{99_appendix}

\end{document}

%% file: colors.tex
\definecolor{LightCyan}{rgb}{0.88,1,1}
\definecolor{chalkblue}{rgb}{0.671, 0.871, 0.902}
\definecolor{chalkpurple}{rgb}{0.796, 0.667,0.796}
\definecolor{chalkyellow}{rgb}{1.0, 1.0, 0.71}
\definecolor{chalkorange}{rgb}{1.0, 0.8, 0.714}
\definecolor{chalkpink}{rgb}{0.953, 0.69,0.765}

\definecolor{bohored}{RGB}{209, 133, 119}
\definecolor{bohogreen}{RGB}{130, 157, 136}
\definecolor{bohoblue}{RGB}{160, 174, 189}
\definecolor{bohoyellow}{RGB}{225, 180, 105}

\definecolor{mellowpurple}{RGB}{213, 104, 171}
\definecolor{mellowpink}{RGB}{250, 200, 220}
\definecolor{mellowgreen}{RGB}{90, 146, 91 }
\definecolor{mellowblue}{RGB}{177, 222, 220}
\definecolor{mellowyellow}{RGB}{ 254, 192, 93}
\definecolor{melloworange}{RGB}{226, 117, 76}

\definecolor{sunsetpink}{RGB}{162, 59, 85}

\definecolor{applegreen}{rgb}{0.55, 0.71, 0.0}
\definecolor{airforceblue}{rgb}{0.36, 0.54, 0.66}
\definecolor{amethyst}{rgb}{0.6, 0.4, 0.8}
\definecolor{antiquefuchsia}{rgb}{0.57, 0.36, 0.51}
\definecolor{aquamarine}{rgb}{0.5, 1.0, 0.83}
\definecolor{asparagus}{rgb}{0.53, 0.66, 0.42}
\definecolor{babyblue}{rgb}{0.54, 0.81, 0.94}
\definecolor{babyblueeyes}{rgb}{0.63, 0.79, 0.95}
\definecolor{babypink}{rgb}{0.96, 0.76, 0.76}
\definecolor{darkseagreen}{rgb}{0.56, 0.74, 0.56}
\definecolor{flavescent}{rgb}{0.97, 0.91, 0.56}
\definecolor{grannysmithapple}{rgb}{0.66, 0.89, 0.63}
\definecolor{pastelorange}{rgb}{1.0, 0.7, 0.28}
\definecolor{pastelmagenta}{rgb}{0.96, 0.6, 0.76}
\definecolor{richelectricblue}{rgb}{0.03, 0.57, 0.82}
\definecolor{rosevale}{rgb}{0.67, 0.31, 0.32}
\definecolor{sandstorm}{rgb}{0.93, 0.84, 0.25}

\definecolor{veryperi}{cmyk}{69,61,0,0}
\definecolor{popcorn}{cmyk}{3,13,53,0}
\definecolor{bubblegum}{cmyk}{3,67,25,0}
\definecolor{orchidbloom}{cmyk}{24,35,4,0}
\definecolor{daffodil}{cmyk}{0,29,76,0}
\definecolor{poinciana}{cmyk}{14,89,92,4}
\definecolor{harborblue}{cmyk}{87,37,44,27}
\definecolor{Cascade}{cmyk}{57,4,36,0}
\definecolor{spunsugar}{cmyk}{33,1,7,0}
\definecolor{coccamocha}{cmyk}{36,48,55,34}
\definecolor{fragilesprout}{cmyk}{34,13,93,1}
\definecolor{supersonic}{cmyk}{95,64,10,1}

%% file: macros.tex
\newcommand{\R}{\mathbb{R}}

\newcommand{\N}{\mathbb{N}}

\newcommand{\D}{\mathcal{D}}
\renewcommand{\L}{\mathcal{L}}

\newcommand{\argmin}{\operatorname{arg\,min}}

\renewcommand{\O}{\mathcal{O}}
\renewcommand{\l}{\ell}



%% file: 00_abstract.tex
Modern machine learning models rely on large amounts of labeled data. 
However, manual annotation of large-scale datasets is expensive and time-consuming. Label spreading is a semi-supervised learning technique that addresses this challenge by propagating information from a few labeled examples to a larger pool of unlabeled data. Despite its effectiveness, its application to large-scale, high-dimensional datasets is limited by computational costs and memory constraints.
To address these limitations, we propose Algebraic Multigrid Acceleration for Efficient Label Spreading (\AMELS), an efficient label spreading framework that improves scalability by fast construction of neighborhood graphs and the incorporation of algebraic multigrid solvers. The latter is an iterative solver that replaces the ordinary random walk iteration typically performed in label spreading. Due to the multilevel nature of algebraic multigrid solvers, \AMELS\ spreads given label information across a graph of any size in a single multigrid cycle.
We demonstrate that \AMELS\ achieves significant runtime reductions compared to existing implementations while also being more robust to hyperparameter choices in terms of both runtime and classification accuracy.
Our framework therefore enables efficient label spreading on large-scale image datasets and produces accurate labels even when only a few labeled samples are available.

%% file: 01_introduction.tex
Large-scale annotated datasets are the foundation of modern machine learning models~\cite{denton2021,Geiger2021}.
Manual annotation of these datasets is time-consuming and costly, as labels must be accurate to ensure effective model training~\cite{Nahum2024LLM,jakubik2024Improve,Song2023LearningFromNoisyLabels}.
The effort becomes even more pronounced for soft labeling approaches, where each data point is annotated multiple times to capture uncertainty or annotator disagreement~\cite{peterson2019human,schmarje2022benchmark}. 
Consequently, obtaining high-quality labeled data represents a major bottleneck for scaling machine learning systems.
To address this challenge, semi-supervised learning methods~\cite{semi-supervised-learning,introduction_to_SSL_2009} leverage the structure of the full dataset to propagate labels from a small set of labeled examples to large amounts of unlabeled data.
In particular, label spreading~\cite{Zhou_2003} constructs a similarity graph from the dataset, assuming that nearby points, or points on the same manifold, are likely to share the same label.
Label predictions are then obtained via an iterative diffusion process on the graph, in which each node exchanges label information with its neighbors until convergence.
The diffusion strength is controlled by a parameter $\alpha \in (0,1)$, where $0$ corresponds to no diffusion and $1$ to maximal diffusion {with minimal preservation} of the initial labels.

{The algorithm} faces two major computational challenges when applied to large-scale datasets:
(1) the construction of a similarity graph, which requires the computation of all pairwise similarities between data points, which scales quadratically with the number of samples, and
(2) the inversion of a matrix to determine the label estimates, scaling cubically in the number of samples and requiring quadratic memory.
These limitations restrict the applicability of label spreading in modern settings~\cite{Iscen2019}, where both dataset size and feature dimension are large.

While several approaches improve the efficiency of label spreading via approximations, sparsification, or iterative solvers, they still do not fully meet the scalability requirements of large-scale settings such as deep learning. A detailed discussion of related work is provided in the corresponding section. Moreover, no widely used efficient implementation of label spreading is currently available.

\begin{figure}[t]
    \centering
    \includegraphics[width=.95\linewidth]{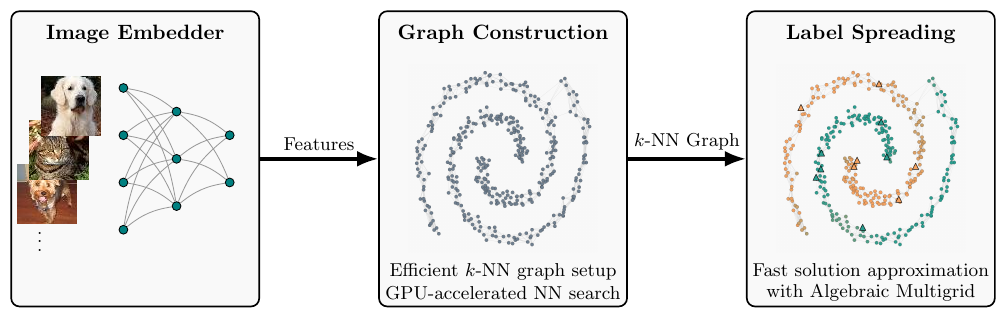}
    \caption{Overview of our proposed framework \AMELS: To apply label spreading on large-scale image datasets, a $k$-NN graph is constructed in the embedding space of a feature extractor such as CLIP. For efficient graph construction, we use FAISS's implementation that accelerates nearest neighbor search on a GPU. Most importantly, to efficiently approximate the solution to label spreading, we employ AMG solvers which iteratively coarsen and interpolate the problem, significantly reducing computational complexity.}
    \label{fig:method}
\end{figure}

\subsubsection*{Our Contribution}
To address the computational challenges in label spreading, we propose Algebraic Multigrid Acceleration for Efficient Label Spreading (\texttt{AMELS}), a framework that:

(1) speeds up graph construction by leveraging semantically meaningful low-dimensional image embeddings obtained from vision-language models (VLMs) and state-of-the-art methods for efficient nearest-neighbor search~\cite{faisslibrary}, to construct a sparse \(k\)-nearest-neighbor graph;

(2) leverages the optimal complexity of algebraic multigrid (AMG) solvers~\cite{trottenberg2000multigrid} to approximate the solution of the linear system involved in label spreading. 
AMG methods are well suited for label spreading, as their hierarchical structure efficiently captures both local and distant graph connections. In particular, coarse levels accelerate the propagation of label information over large distances, which is crucial since such global dependencies are difficult to capture in standard implementations.
Moreover, prior work suggests a diffusion strength $\alpha$ close to $1$ in practice~\cite{klees2026}, resulting in ill-conditioned graph Laplacians that AMG methods can handle robustly.

Our empirical evaluation on standard image classification benchmarks shows that \AMELS significantly reduces the computational cost of label spreading, achieving lower runtime and fewer iterations than existing approaches. We further demonstrate that large-scale image datasets can be labeled accurately at a fraction of the annotation cost. Finally, we present hyperparameter studies on the effects of the parameters of label spreading, showing that \AMELS is robust to parameter choices in terms of both accuracy and runtime.
We make our code publicly available under \url{https://github.com/JonathanKlees/efficient_label_spreading}.
Our main contributions can be summarized as follows:
\begin{itemize}[left=0pt]
    \item We combine efficient neighborhood construction methods and AMG solvers to improve the computational efficiency of label spreading, and leverage VLMs to enable its application to high-dimensional and large-scale image datasets.
    \item We demonstrate that our framework achieves significant efficiency gains over existing approaches while preserving performance in terms of classification accuracy.
    \item We provide a comprehensive hyperparameter study that reveals that our framework is more robust to the choice of hyperparameters regarding both runtime and performance.
    \item We make our efficient label spreading framework publicly available to facilitate future research and applications.
\end{itemize}

%% file: 02_related_work.tex
\subsubsection{Label Propagation and Label Spreading}
Label propagation and label spreading are graph-based semi-supervised methods. Labeled and unlabeled data points are connected in a graph and information from labeled nodes propagates to unlabeled data points.
In \emph{label propagation}~\cite{zhu2003labelprop}, a node's label information is propagated to neighboring nodes according to their similarity, while the labels of the labeled data are kept fixed (hard clamping).
Building on this idea, \emph{label spreading} ~\cite{Zhou_2003} employs a normalized version of the graph Laplacian for regularization and incorporates a soft clamping mechanism, which improves stability and robustness.
Both approaches admit a random walk interpretation and can be formulated as fixed-point iterations. 
Related random walk-based propagation methods are introduced by \cite{Szummer2001} and \cite{Wu2012}.
Label spreading is commonly solved using an iterative update rule proposed by \cite{Zhou_2003}, which computes a truncated Neumann series instead of explicitly inverting the underlying linear system. The implementation in~\cite{scikit-learn} follows this approach.
Each iteration incorporates an additional Neumann term of the Neumann series and updates the solution until convergence.
According to~\cite{efficient_label_propagation}, we refer to this iterative scheme as the \emph{power method}.\footnote{Note that this is not the classical power method for eigenvalues.} While this is a natural approach, it can raise substantial computational cost, particularly for large graphs. 
Consequently, several approaches have been proposed to increase efficiency, which we discuss below.

\subsubsection{Efficient Label Propagation and Label Spreading Methods.}
Adapted implementations address the two main bottlenecks in label spreading: graph construction and determining the label spreading solution. 
\newline
\emph{Graph construction.}
To reduce the computational cost of graph construction, sparse neighborhood graphs are commonly employed. In particular, $k$-nearest neighbors ($k$-NN) graphs are widely used, where each node is connected only to its $k$ closest neighbors~\cite{James2013statisticallearning}. 
This reduces the number of edges from $\O(n^2)$ to $\O(kn)$ compared to fully connected graphs.
As a consequence, the resulting graph Laplacian is also sparse, which significantly reduces memory requirements and the cost of matrix-vector operations involved in label spreading~\cite{scikit-learn,klees2026}.
However, constructing a $k$-NN graph can still be computationally expensive for large-scale datasets, as nearest neighbor search is non-trivial. To address this, our implementation integrates optimized libraries such as $\operatorname{Faiss}$~\cite{faisslibrary} that leverage GPU acceleration to efficiently construct large-scale neighborhood graphs. 
\newline
Beyond standard $k$-NN constructions, recent work proposes efficient construction of similarity graphs using bipartite graph structures, which enable low-rank approximations and iterative refinement based on label smoothness~\cite{Wang2023bipartite,peng2024labelpropbipartit}. While these approaches improve scalability, they do not preserve the full graph structure and may introduce approximation errors. Sparsity can be enhanced by combining multiple graph Laplacians \cite{karasuyma25013_multiplegraphlabelprop}, at the cost of increased model complexity.
\newline
\emph{Determining the label spreading solution.}
Determining the label spreading solution is computationally even more expensive than graph construction. Since computing the inverse of an $n \times n$ system matrix requires $\O(n^3)$, determining the labels for all unlabeled data points in $c$ classes results in computational cost of $\O(n^3 + c n^2)$ \cite{JMLR:v7:belkin06a}.
Fujiwara et al.\ 
\cite{efficient_label_propagation}
reduce the computational cost of the power method to $\O(cnt)$, where $t$ denotes the average number of iterations per label and typically $t \ll n$. Their algorithm accelerates the iteration by removing negligible entries in the series representation of the system matrix, while preserving the exact solution. Nevertheless, no publicly available implementation exists. While this approach reduces computational complexity, it still inherits a key limitation of the {power method}, namely that after \(t\) iterations information can only propagate along paths of length at most \(t\). In contrast, the multilevel hierarchy of our AMG-based framework enables global label propagation within a single iteration. As a result, our framework scales efficiently to large problems and achieves significantly faster convergence. 
In~\cite{Iscen2019}, label spreading is applied in the feature space of a neural network layer to generate pseudo labels for supervised training. Due to the large dataset size, the conjugate gradient (CG) method is applied to solve the linear system arising in label spreading. Similarly, \cite{Fujiwara2021_anchor} apply CG to avoid matrix inversion in label spreading and use $k$-means clustering to construct a graph consisting only of anchor points. We compare our method to optimized CG implementations for both CPU and GPU execution. 
However, while CG degrades significantly for ill-conditioned systems arising in label spreading, our AMG-based framework handles such systems more robustly.
In addition, GPU support in our framework improves runtime compared to publicly available implementations, yielding superior scaling behavior with respect to the system size compared to CG and power methods.

%% file: 03_preliminaries.tex
\subsubsection{Label Spreading.}

Let $\D = \{ x_1, \ldots, x_a, x_{a+1}, \ldots,  x_n \} \subset \R^m$ be a dataset of $n \in \N$ samples,
where the first $a < n$ samples are labeled with labels from $\L = \{1, \ldots, c\}$.
The goal of label spreading is to predict labels for the remaining $n-a$ unlabeled samples~\cite{Zhou_2003}.
To achieve this, pairwise relationships between samples are represented by a similarity graph $G = (V, E, W)$ constructed over the dataset $\mathcal{D}$. Each node $i \in V$ corresponds to a sample $x_i$, while edges $E \subseteq V \times V$ connect pairs of samples according to their similarity.
The edge weights are computed using a Gaussian kernel, i.e.,\begin{equation}\label{eq:gaussiankernel}
W_{ij} =
\begin{cases}
\exp\!\left( \frac{- \lVert x_i - x_j \rVert^2}{2\sigma^2} \right), & \text{if } i \neq j, \\[0.5em]
0, & \text{if } i = j ,
\end{cases}
\end{equation} 
where $\sigma > 0$ is a scaling parameter depending on the underlying structure of the graph \cite{Zhou_2003,bishop2007}.
Let $D\in\mathbb{R}^{n \times n}$ denote the diagonal degree matrix with
\(
D_{ii} = \sum_j W_{ij}.
\)
The weight matrix $W\in\mathbb{R}^{n \times n}$ is then symmetrically normalized
\begin{equation}
 \label{eq:symmetric_normalization_of_W}
     S = D^{-\nicefrac{1}{2}} \cdot W \cdot D^{-\nicefrac{1}{2}}.
\end{equation}
Let \(Y \in \mathbb{R}^{n \times c}\) denote the initial label matrix, where \(Y_{ij}=1\) if \(x_i\) is assigned to class \(j\), and \(Y_{ij}=0\) otherwise. Given a diffusion parameter \(\alpha \in (0,1)\), label spreading propagates initial label information iteratively according to
\begin{equation}
\label{eq:iterative_classification}
    F^{(t+1)} = \alpha S F^{(t)} + (1-\alpha)Y,
\end{equation}
where \(F^{(t)} \in \mathbb{R}^{n \times c}\) denotes the classification matrix at iteration \(t\), with initial condition \(F^{(0)} = Y\).
The first term of \cref{eq:iterative_classification} propagates label information across the graph according to $S$, while the second preserves the initially known labels. The degree of information exchange is steered by $\alpha$. 
 This iteration also admits a random walk interpretation.
  Each entry $F_{i\ell}(t)$ represents the score associated with assigning class $\ell$ to sample $x_i$. 
The iteration converges to the fixed point \mbox{$F^* = (1-\alpha)(I- \alpha S)^{-1} \cdot Y$} \cite{Zhou_2003}. 
 Since the factor $(1- \alpha)$ does not affect label assignment, the classification matrix can equivalently be written as \mbox{$F^* = (I - \alpha S)^{-1}Y$}.
Consequently, label spreading reduces to the sparse linear system 
\begin{equation}\label{eq:labelspread_linearsystems}
    (I- \alpha S)F =Y.
\end{equation}
Since $F, Y \in \mathbb{R}^{n \times c}$, this requires solving $c$ linear systems with the same symmetric positive definite (SPD) system matrix $A:=I - \alpha S$. As these systems scale with the dataset size $n$, efficient iterative solvers are essential in practice.

\subsubsection{Iterative Solvers and Krylov Subspace Methods.}
Krylov subspace methods such as CG and generalized minimal residual method (GMRES) are attractive for solving large sparse linear systems such as~\cref{eq:labelspread_linearsystems} due to their scalability and low memory requirements \cite{saad2013_iterative,trefethen_numerical_2022}.
Consider the linear system $Ax=b$, where \mbox{$A \in \R^{n \times n}$} is SPD.
{A basic iterative scheme is
\begin{equation}\label{eq:basic_iterative_sheme}
    x^{(t +1)} = x^{(t)} + B(b - Ax^{(t)})
\end{equation}
where $B \approx A ^{-1}$.  
}
Given an iterate $x^{(t)}$, residual and error are
\begin{equation}\label{eq:residual_and_error}
    r^{(k)} = b - A x^{(k)}, 
\qquad 
e^{(t)} = x^\star - x^{(t)} = A^{-1} r^{(t)},
\end{equation}
where $x^\star$ denotes the exact solution. 
The error propagates as 
\begin{equation*}
    e^{(t +1)} = (I- BA)e^{(t)}.
\end{equation*}
Hence, convergence depends on the iteration matrix $I-BA$ and the method converges rapidly if $BA \approx I$. Such iterative methods quickly dampen the high-frequency error components,
a process commonly referred to as \textit{smoothing}.
Given the initial residual $r^{(0)} = b - Ax^{(0)}$, Krylov methods construct iterates in the $t$-dimensional Krylov subspace
\begin{equation*}
   \mathcal{K}_{t}(A, r^{(0)}) := \operatorname{span}(r^{(0)}, Ar^{(0)}, A^2r^{(0)}, \ldots, A^{t - 1} r^{(0)}).
\end{equation*}
At iteration $t$, a Krylov method computes the best approximation solution in $K_t$, according to a method-specific criterion.
For SPD matrices, CG minimizes the error in the $A$-norm, 
\begin{equation*}
    || x ||_A := \sqrt{x^T Ax}.
\end{equation*}
More generally, GMRES computes iterates that minimize the Euclidean residual norm over the affine subspace
$K_t$~\cite{trefethen_numerical_2022}.

Since \(r^{(k)} = A e^{(k)}\) by~\cref{eq:residual_and_error}, it follows that
\begin{equation}\label{eq:residual_and_error_bound}
    \|e^{(k)}\|_2 \le \|A^{-1}\|_2 \, \|r^{(k)}\|_2 .
\end{equation}
Thus, despite the favorable residual minimization properties, convergence depends strongly on the spectral properties of \(A\). Convergence may deteriorate for ill-conditioned systems with large \(\|A^{-1}\|_2\), such that a small residual does not necessarily imply a small error.
For SPD matrices, 
\begin{equation}
     \|A^{-1}\|_2 = \frac{1}{\lambda_{\min}(A)},
\end{equation}
with $\lambda_{\min}(A)$ denoting smallest eigenvalue of $A$. Hence, eigenvalues near zero may separate residual and error.

\subsubsection{Algebraic Multigrid (AMG) Methods.}
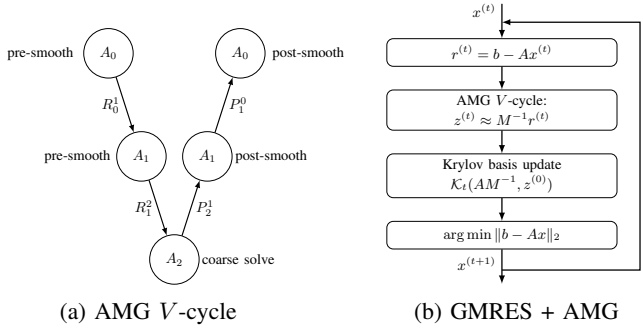
\begin{figure}
\begin{subfigure}[t]{.45\linewidth}
    \centering
    \resizebox{1.2\linewidth}{!}{
        \input{tikz/V-cycle}
    }
    \caption{AMG $V$-cycle}\label{subfig:V-cycle}
\end{subfigure}%
\hfill
\begin{subfigure}[t]{.45\linewidth}
\centering
    \resizebox{0.9\linewidth}{!}{
        \input{tikz/gmres+v-cycle2}
    }
    \caption{GMRES + AMG}\label{subfig:gmres+amg}
\end{subfigure}
\caption{(a): AMG $V$-cycle with $2$ levels. On each level pre- and post-smoothing steps based on \cref{eq:basic_iterative_sheme} are applied, while the coarsest level is solved directly. (b) Schematic illustration of GMRES with AMG preconditioning.}\label{fig:V-cylce_gmres-amg-precond}
\end{figure}

To improve spectral properties and convergence, preconditioning is commonly applied, resulting in the equivalent system
\begin{equation}
     A M^{-1}z =b, \quad x = M^{-1}z, 
\end{equation}
where $M^{-1} \approx A^{-1}$ denotes a suitable preconditioner (earlier denoted as $B$). In this work, AMG is used to approximate the action of $M^{-1}$.  
AMG provides an efficient hierarchical framework for solving large systems by combining local smoothing with coarse-grid corrections~\cite{trottenberg2000multigrid}.

AMG consists of a \textit{setup phase} and a \textit{solve phase}. In the \textit{setup phase}, a hierarchy of coarser systems is constructed,
\begin{equation*}
    A_0x_0 = b_0, \; A_1x_1 = b_1,\, \ldots, \; A_Lx_L = b_L,
\end{equation*}
where $A_0=A$ and coarse operators are recursively defined: 
\begin{equation*}
    A_{\l+1} = R_\l^{\l+1} A_\l P_{\l+1}^\l.
\end{equation*}
Here, $P_{\l+1}^\l$ and $R_\l^{\l+1}$ are prolongation and restriction operators between adjacent levels, where typically \mbox{$R_\l^{\l+1} = (P_{\l+1}^\l)^T$} is chosen.
The coarsening and interpolation operators 
are determined by selecting coarse and fine variables, i.e., aggregating strongly connected variables into coarser representations.

In the \textit{solve phase}, the hierarchy is applied recursively. Starting on a fine level $\l$, pre-smoothing, e.g.\ \cref{eq:basic_iterative_sheme}, is applied to reduce high-frequency error components. The residual is restricted to a coarser level $\l+1$, where low-frequency components appear as high-frequency components and can be reduced more efficiently. This process is repeated until the coarsest level $L$ is reached where the coarse system is solved directly. The resulting solution is subsequently interpolated back to the finer levels to correct the fine-grid solutions, followed by post-smoothing, e.g.~\cref{eq:basic_iterative_sheme}. The combination of smoothing and coarse-grid correction forms an AMG $V$-cycle (cf.~\cref{subfig:V-cycle}) and reduces both high- and low-frequency error components efficiently.
For large sparse systems, this multilevel structure often yields near-linear complexity~\cite{trottenberg2000multigrid}.\\ 
\textit{GMRES with AMG preconditioning.}
In every GMRES iteration, one AMG $V$-cycle applies the preconditioner $M^{-1}\approx A^{-1}$ to the current residual, computing
\begin{equation}
    z^{(t)} \approx M^{-1}r^{(t)}.
\end{equation}
Let $z^{(0)}  = M^{-1}r^{(0)}$ denote the initial preconditioned residual. GMRES then constructs iterates in the Krylov subspace 
\begin{equation}
    \mathcal{K}_t(AM^{-1},z^{(0)}),
\end{equation}
At iteration $t+1$, GMRES minimizes the residual norm over the affine Krylov subspace~\cite{saad_schultz1986} 
\begin{equation}
    x^{(t +1)}  = \underset{x \in x^{(0)} + M^{-1}\mathcal{K}_t}{\argmin} \| b - Ax\|_2.
\end{equation}
A schematic illustration of this process is shown in~\cref{subfig:gmres+amg}.

%% file: tikz/V-cycle.tex
\begin{tikzpicture}[
    level/.style={circle, draw, 
    minimum size=12mm},
    arrow/.style={-latex, thick},
    scale=1
]

\node[level] (A2) at (0,0) {$A_2$};
\node[level] (A1l) at (-.8, 2.5) {$A_1$};
\node[level] (A1r) at (.8, 2.5) {$A_1$};
\node[level] (A0r) at (1.6, 5) {$A_0$};
\node[level] (A0l) at (-1.6, 5) {$A_0$};

\draw[arrow] (A0l) -- node[left] {$R_0^1$} (A1l);
\draw[arrow] (A1l) -- node[left] {$R_1^2$} (A2);

\draw[arrow] (A2) -- node[right] {$P_2^1$} (A1r);
\draw[arrow] (A1r) -- node[right] {$P_1^0$} (A0r);
;

 \node at (1.55, 0) {coarse solve};
\node at (-3.25, 5) {pre-smooth};
\node at (-2.35, 2.5) {pre-smooth};

\node at (3.25, 5) {post-smooth};
\node at (2.35, 2.5) {post-smooth};
\end{tikzpicture}

%% file: tikz/gmres+v-cycle2.tex
\begin{tikzpicture}[
    node distance=5mm,
    block/.style={
        draw,
        rounded corners,
        align=center,
        minimum width=50mm,
        minimum height=6mm
    },
    arrow/.style={-latex, thick},
    every node/.style={font=\small}
]

\node[block] (res) {$r^{(t)} = b - A x^{(t)}$};

\node[block, below=of res] (amg)
{AMG $V$-cycle:\\[1.75pt]
$ z^{(t)} \approx M^{-1} r^{(t)}$};

\node[block, below=of amg] (z)
{
Krylov basis update\\[1.75pt]
$\mathcal K_t(AM^{-1},z^{(0)})$};

\node[block, below=of z](xnext)
{
$\argmin
\|b-Ax\|_2$};



\draw[arrow] (res) -- (amg);
\draw[arrow] (amg) -- (z);
\draw[arrow] (z) -- (xnext);

\draw[arrow] ($(res.north)+(0,0.75)$) -- node[pos=0.2,left] {$x^{(t)}$} (res.north);

\draw[arrow] (xnext.south)
-- node[pos=0.4,left] {$x^{(t+1)}$} ++(0,-0.75)  coordinate (out); 

\path (xnext.south) -- (out) coordinate[pos=0.6] (midout);

\draw[arrow] (midout) -- ++(3,0) |- ($(res.north)+(0,0.35)$);

\end{tikzpicture}

%% file: 04_method.tex
In this section, we present our framework, Algebraic Multigrid Acceleration for Efficient Label Spreading (\AMELS), which addresses the key computational challenges associated with label spreading, namely, graph construction and determining the label spreading solution.
An overview of our proposed framework is depicted in~\cref{fig:method}. In the following, we provide a detailed description of how we address these challenges.

\subsection{Graph Construction}
\subsubsection{Embedding.}
Semi-supervised approaches commonly rely on the assumption of local smoothness~\cite{Zhou_2003}, which states that samples that are close in the Euclidean norm are likely to share the same semantic label.
Constructing an informative graph therefore depends on how well distances reflect similarity relationships in the data.
However, in high-dimensional spaces this requirement is often violated due to the curse of dimensionality~\cite{bellmann1957}, as distance measures lose discriminative power. 
To address this issue for high-dimensional inputs such as images, we first apply a feature extraction step that maps the data into a lower-dimensional representation. 
In this reduced space, distances become more informative and the local smoothness assumption is more plausibly satisfied.
While our framework is agnostic to the choice of any dimensionality reduction technique, a robust feature extraction pipeline proposed by~\cite{klees2026} uses CLIP~\cite{CLIP} for image feature extraction followed by UMAP~\cite{UMAP} for further dimensionality reduction.
\subsubsection{Nearest neighbors.}
Following~\cite{Zhou_2003}, we build the affinity matrix $W$ using a Gaussian kernel, cf.~\cref{eq:gaussiankernel}, to quantify pairwise similarities in the embedding space of the image dataset.
To ensure graph sparsity, as in~\cite{Iscen2019}, an edge between two nodes is introduced only if at least one node belongs to the $k$-NN of the other. 
In contrast to fully connected graphs,  $k$-NN graphs can be constructed efficiently for large datasets~\cite{Iscen2017Efficient}.
The resulting sparse matrix $W$ is then symmetrized and subsequently normalized using the degree matrix $D$ according to~\cref{eq:symmetric_normalization_of_W}. 
The Gaussian kernel bandwidth parameter $\sigma^2$ is chosen as the mean squared $k$-NN radius, averaged over all data points.
We use the Faiss library~\cite{faisslibrary} to efficiently construct $k$-NN graphs. 
Its GPU-optimized implementation enables parallel distance computations, which significantly accelerates nearest neighbor search  for large-scale datasets.

\subsection{Efficient Solution Approximation via AMG}
The key contribution of our framework is leveraging the hierarchical structure inherent to AMG methods and effective preconditioning to efficiently approximate the solution of the label spreading system~\cref{eq:labelspread_linearsystems}.
 As shown in~\cref{sec: Preliminaries}, convergence of iterative solvers depends on the spectrum of the system matrix.
For label spreading, the system matrix is 
\begin{equation*}
    A = I - \alpha S.
\end{equation*}
If $\mu_i$ denote the eigenvalues of $S$, 
the eigenvalues of $A$ are
\[
\lambda_i = 1 - \alpha \mu_i.
\]
In particular, $S$ has the largest eigenvalue $\mu_{\mathit{max}} = 1$ with eigenvector $D^{\nicefrac{1}{2}}\mathbf{1}$, as
\begin{equation*}
    S D^{\nicefrac{1}{2}}\mathbf{1} = D^{-\nicefrac{1}{2}} W \mathbf{1} = D^{\nicefrac{1}{2}}\mathbf{1},
\end{equation*}
where $W \mathbf{1} = D \mathbf{1}$ by definition of the degree matrix. Hence,
\begin{equation}
\label{eq: smallest eigenvalue}
\lambda_{\min} = 1 - \alpha,
\qquad
\|A^{-1}\|_2 = \frac{1}{\lambda_{\min}} = \frac{1}{1-\alpha}.
\end{equation}
In practice, large-scale datasets with limited annotation budgets require a diffusion strength \(\alpha\) close to 1 to ensure sufficient label propagation. Moreover, label spreading on neighborhood graphs is consistent in the large-data limit when the regularization increases with dataset size, i.e., \(\alpha \to 1\) as \(n \to \infty\)~\cite{klees2026}. In this setting, the smallest eigenvalue of the system approaches zero and \(\|A^{-1}\|_2\) diverges. Consequently, residual-based stopping criteria may severely underestimate the true error by~\cref{eq:residual_and_error_bound}, and Krylov methods can converge slowly without appropriate preconditioning.
To address this issue, we combine GMRES with AMG preconditioning. 
Preconditioning improves the system matrix’s spectral properties by shifting its spectrum away from zero. This increases the effective smallest eigenvalue, improves residual-error alignment, and accelerates convergence.
Consequently, our framework overcomes a key limitation of existing label spreading implementations, namely 
the increasingly ill-conditioned system matrix arising for large diffusion strengths $\alpha$.

To compute the label spreading solution according to~\cref{eq:labelspread_linearsystems},
we solve the class-wise systems
\begin{equation*}
  F^*_{j} = (I - \alpha S)^{-1}Y_{j} \quad \forall\, j\in\{1,\ldots, c\},
\end{equation*}
where $F^*_{j}$ and $Y_{j}$ denote the $j$-th columns of the respective matrices. 
Since all $c$ class score vectors correspond to linear systems with coefficient matrix $I - \alpha S$, the AMG hierarchy is constructed only once during setup and subsequently reused for all right-hand sides.
When solving the resulting linear systems with AMG, the hierarchical structure effectively reduces the number of unknowns on coarser levels, thereby significantly decreasing the computational complexity. Moreover, coarse-grid correction enables global propagation of label information across the graph within a single iteration, in contrast to the {power method} that propagates information only locally in one iteration by~\cref{eq:iterative_classification}.

%% file: 05_results.tex

\begin{figure*}[tb]
\centering

\begin{subfigure}[t]{0.49\linewidth}
    \centering
    \includegraphics[width=0.8\linewidth]{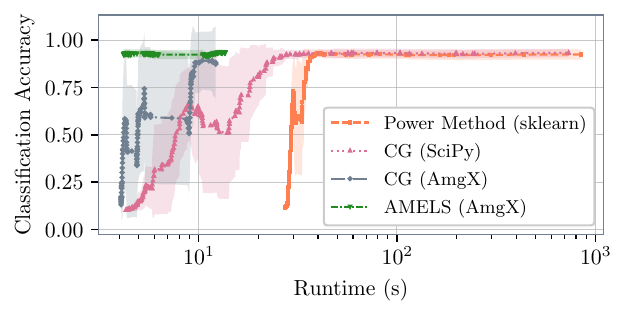}
    \caption{EMNIST-digits}
    \label{fig:performance_by_runtime}
\end{subfigure}
\hfill
\begin{subfigure}[t]{0.49\linewidth}
    \centering
    \includegraphics[width=0.8\linewidth]{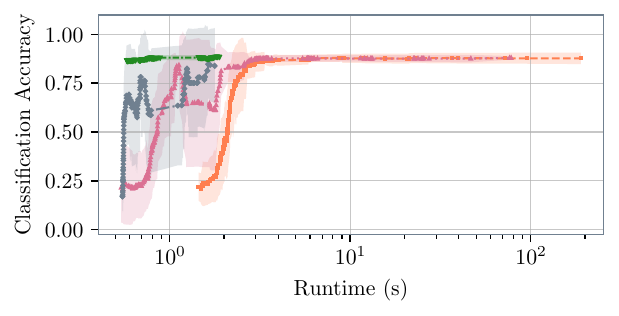}
    \caption{CIFAR-10} 
    \label{fig:CIF_performance_by_runtime}
\end{subfigure}
\hfill
\begin{subfigure}[t]{0.49\linewidth}
    \centering
    \includegraphics[width=0.8\linewidth]{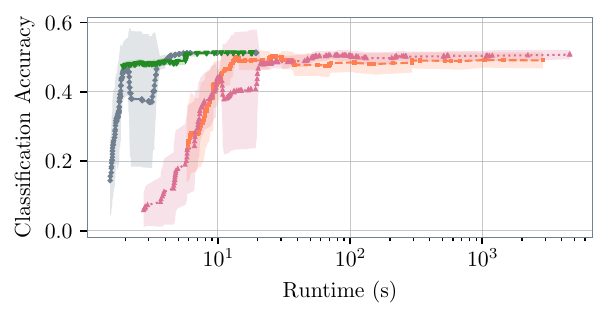}
    \caption{Tiny ImageNet} 
    \label{fig:TIN_performance_by_runtime}
\end{subfigure}
\caption{Classification accuracy of different label spreading implementations on EMNIST-digits, CIFAR-10 and Tiny ImageNet depending on their runtime to approximate the solution. Performance is smoothed with a moving average kernel and empirical standard deviation is displayed.}
\label{fig:CIF&TIN_performance_by_runtime}
\end{figure*}

In this section, we empirically evaluate our approach to answer the following research questions:
\begin{enumerate}[left=0pt]
    \item \textbf{Computational Efficiency:} 
    To what extent does $\AMELS$ improve the computational efficiency of label spreading in comparison to existing baselines?
    \item \textbf{Dependence on Hyperparameters:} 
    How do key hyperparameters influence runtime of $\AMELS$, compared to other label spreading implementations?
     \item \textbf{Classification Accuracy:} 
    Does increased runtime efficiency affect classification accuracy of label spreading?
\end{enumerate}
Before addressing these research questions, we describe our experimental setup. 

\subsection{Experimental Setup}
\label{subsec: Numerical_Results-Experimental_Setup}

\subsubsection{Implementation Details.}
 \AMELS combines Flexible GMRES (FGMRES) as an iterative solver with an AMG preconditioner, where one $V$-cycle iteration is conducted and an initial Krylov subspace of dimension $t =20$ is constructed. 
To determine coarse and fine variables in AMG from the connectivity structure of the system matrix, we use the Parallel Modified Independent Set algorithm~\cite{PMIS} with $\beta=0.25$. The interpolation scheme follows the \emph{extended+$i$} method~\cite{Grother1995} and one post-smoothing step is applied. To solve the linear system at the coarsest level, a dense LU solver is used. 

\subsubsection{Baselines.}
We compare our method against the following baselines:
 \begin{itemize}[left=0pt]
     \item \emph{Power method} in $\operatorname{scikit-learn}$ \cite{scikit-learn} 
     \item \emph{Direct Solve:} sparse direct solve via SciPy $\operatorname{spsolve}$~\cite{scipy}
     \item \emph{CG (SciPy):} iterative solving with the Conjugate Gradient method as implemented in SciPy~\cite{scipy}
     \item \emph{CG (AmgX):} iterative solving with the CG method as implemented in  the AmgX library~\cite{AMGX}
 \end{itemize}
All methods solve the linear system up to the same tolerance for the relative residual norm. For the power method, we modified the stopping criterion accordingly. Initially, convergence was defined by iterates changing only marginally.
Instead, we require the {residual norm to be sufficiently small}, ensuring comparable accuracy across methods. This requires an additional matrix vector multiplication per iteration to compute the residual, which is negligible compared to the overall runtime.

\subsubsection{Datasets.}

We evaluate our method on three image classification datasets of varying scale and complexity. The \mbox{EMNIST-Digits} dataset~\cite{EMNIST} contains $280,\!000$ grayscale images of handwritten digits. Furthermore, we consider the image datasets CIFAR-10~\cite{cifar10}, which consists of $50,\!000$ images from $10$ classes, and Tiny ImageNet~\cite{le2015tiny}, comprising $100,\!000$ images from $200$ classes derived from ImageNet~\cite{ILSVRC15}.

\subsubsection{Hardware Specifications.}
The experiments were conducted using an NVIDIA Quadro P6000 GPU with $24$ GB of memory and two Intel(R) Xeon(R) Gold $6138$ CPUs with $2.00$ GHz and $20$ cores/$40$ threads with $502$ GB of RAM.
Consequently, CPU-based implementations using SciPy~\cite{scipy} rely on two 20-core CPUs, while AMG implementations using the AmgX library~\cite{AMGX} use a single GPU.

\subsubsection{Parameter Configurations.}
Due to the practical relevance of large spreading intensities, we evaluate implementations for a fixed parameter configuration of $k=20$ neighbors and a spreading intensity of $\alpha = 0.99$, providing the algorithm with $100$ randomly drawn labels and specifying a residual tolerance of $10^{-3}$. We construct the neighborhood graph in the embedding space obtained by reducing CLIP ViT-B/32~\cite{CLIP} features of the original images using UMAP~\cite{UMAP} with default parameters and a target dimension of $20$.

\begin{figure*}[tb]
    \centering
    \begin{subfigure}[b]{0.48\textwidth}
        \centering
        \includegraphics[width=0.9\textwidth]{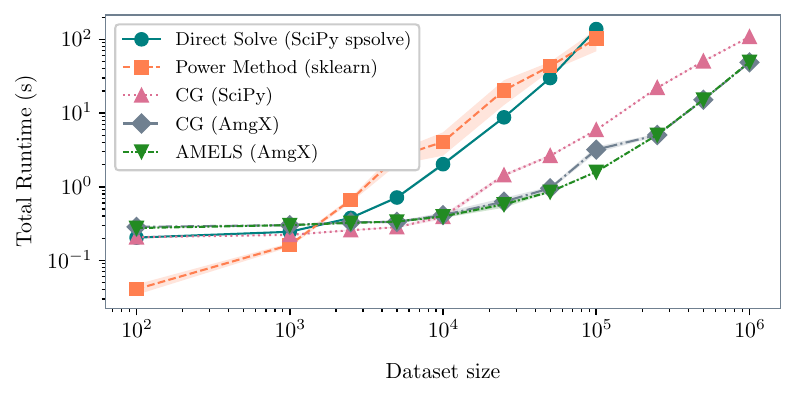}
        \caption{Runtime in seconds.}
        \label{fig:benchmark_comparison_runtime}
    \end{subfigure}
    \hfill
    \begin{subfigure}[b]{0.48\textwidth}
        \centering
        \includegraphics[width=0.9\textwidth]{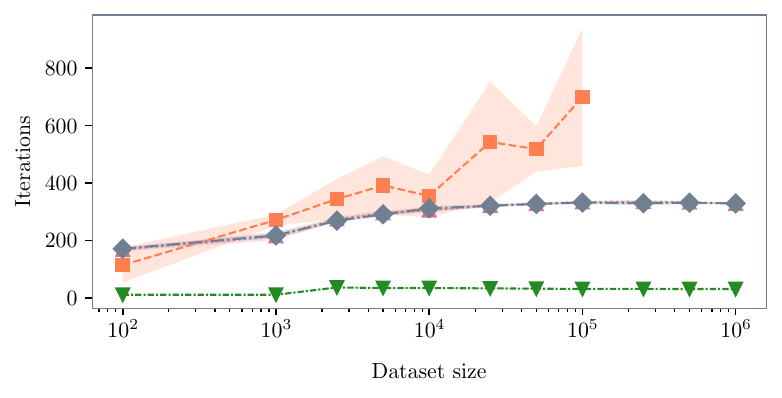}
        \caption{Number of iterations.}
        \label{fig:benchmark_comparison_iterations}
    \end{subfigure}
    \caption{Efficiency comparison of different label spreading implementations. Runtime in seconds (a) and the number of iterations (b) averaged over 10 runs is displayed for varying dataset sizes of the EMNIST-digits dataset.}
    \label{fig:benchmark_comparison}
\end{figure*}

\subsection{Computational Efficiency}
\subsubsection{Classification Accuracy versus Runtime.}
\label{subsec: Numerical_Results-Comparison of Implementation Efficiency}
To compare the efficiency of different label spreading implementations, we evaluate classification accuracy versus runtime.
For this experiment, we vary the maximal number of iterations and measure the resulting classification accuracy.
The number of iterations, runtime, and performance depend on the randomly chosen initial labels. Hence, we smooth the resulting performance using a {moving-average kernel} with a window size of $30$ and compute the empirical standard deviation. \cref{fig:CIF&TIN_performance_by_runtime}
 displays our results. 
 It can be observed that the {power method} requires substantially longer runtimes than competing methods to achieve high classification accuracy. Both CG implementations achieve comparable classification performance at lower computational costs but exhibit high variance. In contrast, \AMELS\ consistently achieves high classification accuracy with minimal runtime, particularly on EMNIST-digits and \mbox{CIFAR-10}.

\begin{figure*}[tb]
\centering

\begin{subfigure}[t]{0.48\linewidth}
    \centering
    \includegraphics[width=\linewidth]{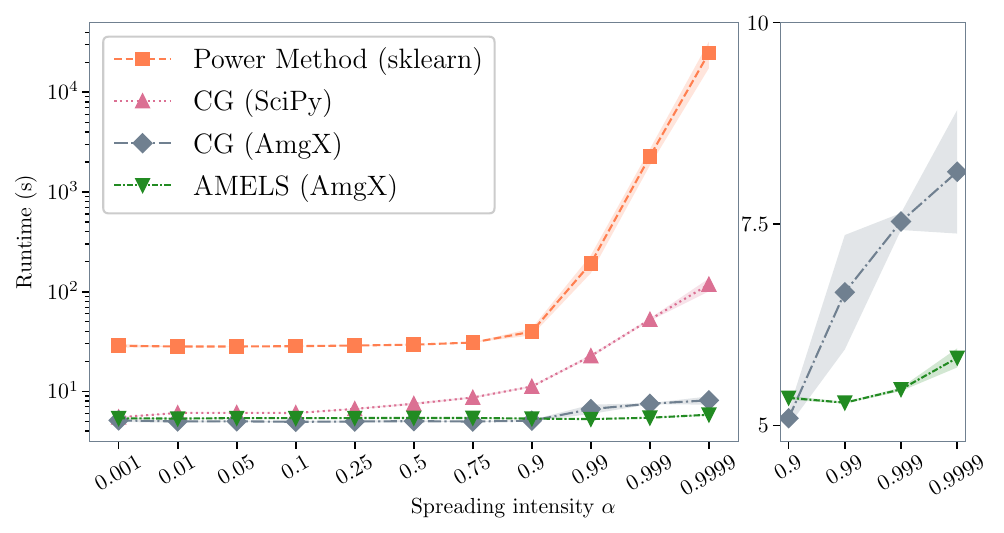}
        \caption{Runtime in seconds.}
        \label{fig:runtime_depending_on_alpha}
\end{subfigure}
\hfill
\begin{subfigure}[t]{0.48\linewidth}
    \centering
    \includegraphics[width=\linewidth]{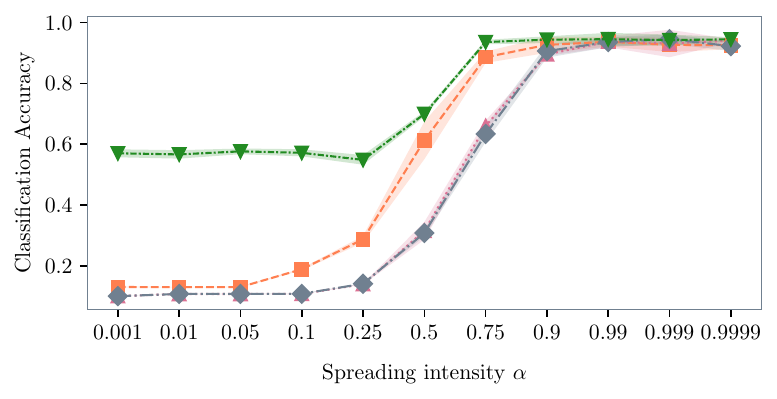}
    \caption{Classification accuracy.}
    \label{fig:accuracy_depending_on_alpha}
\end{subfigure}
\caption{Runtime (a) and classification accuracy (b) of different label spreading implementations on EMNIST-digits depending on the spreading intensity $\alpha$. Our method \AMELS is least sensitive to the choice of $\alpha$ regarding both runtime and performance.}
\label{fig:runtime_and_performance_depending_on_alpha}
\end{figure*}

\subsubsection{Scalability Across Dataset Sizes.}
\label{subsec: Numerical_Results-Scalability}
To investigate scalability, we vary the dataset size and evaluate runtime and the number of iterations until convergence in \cref{fig:benchmark_comparison}. For this experiment, we applied random augmentations to images of the EMNIST-digits dataset to increase its size to $1,\!000,\!000$ data points.
While the {power method} is fastest for very small datasets, both direct solving and the {power method} become computationally costly already for moderately sized datasets. Iterative solving using CG scales better in terms of runtime but still falls short of the runtime efficiency of \AMELS. The number of iterations is not directly comparable between the {power method} and iterative solvers. One iteration in the {power method} corresponds to an update of the estimated system matrix by adding an additional term of its series representation. In contrast, iterative solvers estimate the solution of the linear system class-wise, so that iterations scale with the number of classes. Despite this scaling, the overall number of iterations needed is significantly lower for iterative solvers, particularly for $\AMELS$.

\subsection{Impact of Hyperparameters on Runtime}
\label{subsec: Numerical_Results-Hyperparameters_Runtime}

\subsubsection{Diffusion strength.}
A larger diffusion strength $\alpha$ increases the condition number of the system, making it more difficult to solve. Consequently, the runtime of label spreading methods is expected to increase with larger values of $\alpha$.
This effect is shown in \cref{fig:runtime_depending_on_alpha}.
We excluded direct solving from this experiment due to runtime limitations. The runtime of the {power method} increases sharply as $\alpha \rightarrow 1$, while the SciPy CG implementation also converges substantially slower. In contrast, the runtime of \AMELS\ is only minimally affected, presumably due to preconditioning improving the spectral properties of the system matrix.
Hence, for $\alpha$ close to $1$, AMG-based solving can significantly reduce runtime, also compared to the AmgX implementation of CG.
Apart from preconditioning, another explanation is that label information propagates farther across the graph for large $\alpha$. Accurately capturing these weak long-range interactions requires many iterations for conventional iterative solvers, whereas AMG can approximate them efficiently through its multilevel structure.
\subsubsection{Graph density.}
The number of neighbors $k$ controls the sparsity of the underlying graph. In particular, connecting more neighbors yields a denser matrix, increasing the complexity of solving the linear system. Moreover, the graph construction via nearest neighbor search becomes more computationally expensive.
Hence, $k$ affects the runtime of label spreading in both graph construction and system solving. As shown in~\cref{fig:k_vs_runtime}, both $\AMELS$ and the baseline methods exhibit limited sensitivity to graph connectivity, enabling the use of moderately dense graphs without significant runtime increases.

\subsection{Classification Accuracy}
\label{subsec: Numerical_Results-Hyperparameters_Performance}

\begin{figure}[t]
\centering
\begin{subfigure}[t]{0.48\linewidth}
    \centering
    \includegraphics[width=\linewidth]{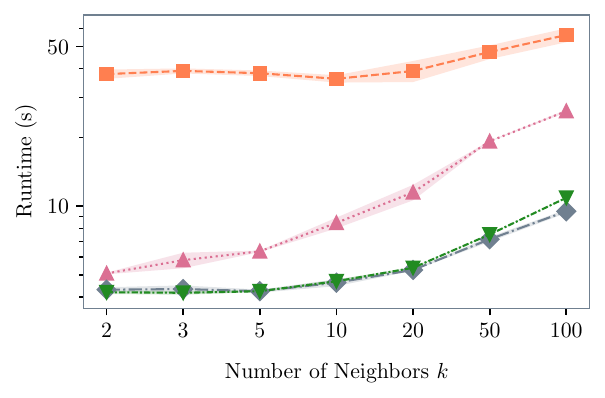}
    \caption{Runtime in seconds.}
    \label{fig:k_vs_runtime}
\end{subfigure}
\hfill
\begin{subfigure}[t]{0.48\linewidth}
    \centering
    \includegraphics[width=\linewidth]{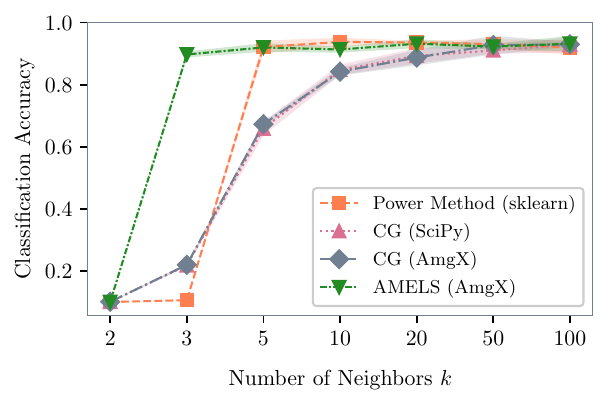}
    \caption{Classification accuracy.}
    \label{fig:accuracy_depending_on_k}
\end{subfigure}
\caption{Runtime (a) and classification accuracy (b) of different label spreading implementations on EMNIST-digits depending on the number of neighbors $k$ used for graph construction.}
\end{figure}

The previous experimental results show that $\AMELS$ significantly reduces the runtime of label spreading. A natural question is whether these runtime savings come at the cost of reduced classification accuracy.
\cref{fig:accuracy_depending_on_alpha} shows that rather the opposite is true.
Performance is best for $\alpha$ close to one and poor for small $\alpha$. However, the degradation is less severe for the {power method} and even less for \AMELS. 
Consequently, with respect to the choice of $\alpha$, \AMELS\ is more robust in terms of achieved classification accuracy.
Similarly, increased robustness can be observed regarding the choice of the hyperparameter $k$. Figure \ref{fig:accuracy_depending_on_k} shows that $\AMELS$ achieves better classification results than competing implementations for a very small number of neighbors.
Overall, it can be concluded that $\AMELS$ boosts the computational efficiency of label spreading without sacrificing performance.

%% file: 06_conclusion.tex
$\AMELS$ is a framework for efficient label spreading that combines sparse neighborhood graphs with AMG methods to improve computational efficiency. 
From a numerical perspective, the usage of AMG methods as preconditioners is desired as it improves the spectral properties of the system matrix involved in label spreading.
Comprehensive numerical experiments show that our framework significantly speeds up label spreading compared to existing implementations.
We demonstrated that this speedup does not come with sacrifices in performance but rather the opposite is true and \AMELS proves to be more robust to the choice of hyperparameters of label spreading.
As a result, $\AMELS$ makes label spreading applicable to large-scale image datasets and enables their semi-automated annotation to reduce the cost of manual labeling.

%% file: 99_appendix.tex
In the appendix, we  provide additional details for the CG method and GMRES and elaborate on the spectral properties of the system matrix involved in label spreading.
%
%
 \subsection{CG and GMRES}

The conjugate gradient (CG) method 
is applicable when $A$ is symmetric positive definite, i.e., $A^T = A$ and $x^T Ax >0 \;  $ for all $x \in \R^n \backslash \{0\}$. The $A$-norm is defined by \begin{equation*}
    || x ||_A := \sqrt{x^T Ax}.
\end{equation*}
The CG method generate iterates $\{x^{(\kappa)} \in x^{(0)} + K^{(\kappa)}(A, r^{(0)})\}$ such that the error is minimized in the $A$ norm, which is equivalent to minimizing
\begin{equation*}
    \phi(x) = \nicefrac{1}{2} \; x^T A  -x^T b,
\end{equation*}
cf.\ \cite{trefethen_numerical_2022}. 
The generalized minimal residual method (GMRES) computes iterates in the affine space $x^{(0)} + K^{(\kappa)}(A, r^{(0)})\}$ by minimizing the Euclidean norm ot the residual, i.e., 
\begin{equation*}
    x^{(\kappa)} = \underset{x \in x^{(0)} + K^{(\kappa)}(A, r^{(0)})}{\argmin} \;\;  || b - A x ||_2
\end{equation*}

Typically GMRES uses the Arnoldi iteration \cite{saad2013_iterative} to create an orthonormal basis $V^{(\kappa)} = \{ \nu_1, \ldots, \nu_\kappa\}$ of $K^{(\kappa)} (A, \nu_1)$, where
\begin{equation*}
\nu_1 = \frac{r^{(0)}}{|| r^{(0)}||_2}.
\end{equation*}
The GMRES iterate is then given by 
\begin{equation*}
    x^{(\kappa)} = x^ {(0)} + V^{(\kappa)} y^{(\kappa)},
\end{equation*}
where $y^{(\kappa)}$ solves the small least-squares problem involving  the corresponding Hessenberg matrix $H^{(\kappa)}$ $$y^{(\kappa)} = \underset{y \in \R^\kappa}{\argmin}  || \beta e_1 - H^{(\kappa)} y||_2 \;\; \text{with  } \beta = || r^{(0)}||_2.$$

\subsection{Spectral Properties of the Graph Laplacian}
\label{sec-appdx:condition_number}
We note the following for the spectral properties of the graph Laplacian. The Laplacian $I-\alpha S$ is a positive definite matrix because $S$ is stochastic, i.e., the rows of $S$ sum up to 1. Consequently, the eigenvalues of $S$
are likewise always less than or equal to 1. In particular, the largest eigenvalue of $S$ is $\mu_{\mathit{max}} = 1$ and the smallest eigenvalue is $\mu_{\mathit{min}} = -1$. This is due to the fact that $D^{\nicefrac{1}{2}}\mathbf{1}$ is an eigenvector of $S$ with eigenvalue 1 as
\begin{align*}
    S D^{\nicefrac{1}{2}}\mathbf{1} &= D^{-\nicefrac{1}{2}} W D^{-\nicefrac{1}{2}} D^{\nicefrac{1}{2}}\mathbf{1} \\
    & = D^{-\nicefrac{1}{2}} W \mathbf{1} \\
    & = D^{-\nicefrac{1}{2}} D \mathbf{1} \\
    & = D^{\nicefrac{1}{2}}\mathbf{1},
\end{align*}
where $W \mathbf{1} = D \mathbf{1}$ by definition of the degree matrix:
\[
(W \mathbf{1})_i = \sum_{j=1}^n W_{i,j},\qquad (D \mathbf{1})_i = D_{ii} = \sum_{j=1}^n W_{i,j} .
\]

This implies that the eigenvalues of $\alpha S$ lie within the interval $[-\alpha, \alpha]$, and consequently, the eigenvalues of
$I-\alpha S$ reside in the interval $[1-\alpha, 1+\alpha]$ where the boundaries are attained by the minimal and maximal eigenvalue. Given that $\alpha \in(0,1)$, it follows that the graph Laplacian is positive definite.

\begin{figure}[tb]
    \centering
    \includegraphics[width=\linewidth]{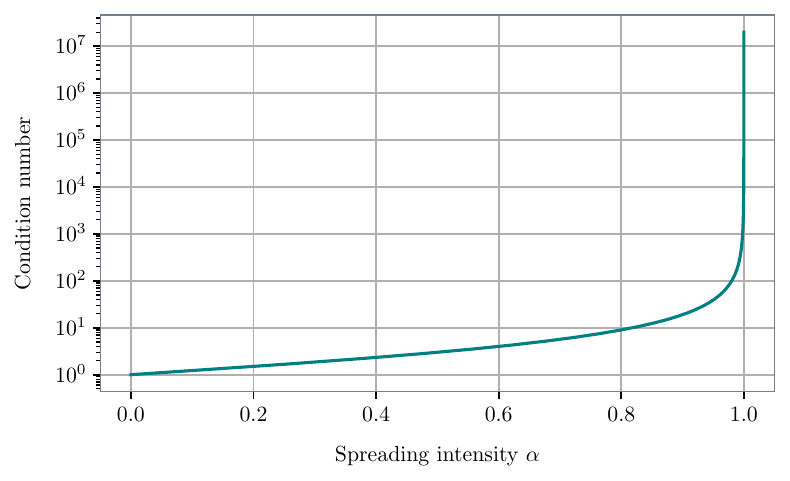}
    \caption{Condition number of the graph Laplacian $I-\alpha S$ depending on the spreading intensity $\alpha$.}
    \label{fig: condition_number_by_formula}
\end{figure}

The condition number of a matrix is defined by the ratio of its largest and its smallest eigenvalue.
For the graph Laplacian $I-\alpha S$, the smallest eigenvalue is given by $\lambda_{\mathit{min}} = 1-\alpha$ and the largest eigenvalue is given by $\lambda_{\mathit{max}} = 1+\alpha$. Hence, the condition number is given as
\begin{equation}
    \label{eq: condition_number_graph_laplacian}
    \kappa(I-\alpha S) = \frac{1+\alpha}{1-\alpha}.
\end{equation}
In \autoref{fig: condition_number_by_formula}, the condition number of the graph Laplacian $I-\alpha S$ is displayed for varying spreading intensity $\alpha$.

